# Correlation Flow: Robust Optical Flow Using Kernel Cross-Correlators

Chen Wang, Tete Ji, Thien-Minh Nguyen, and Lihua Xie

*Abstract*— Robust velocity and position estimation is crucial for autonomous robot navigation. The optical flow based methods for autonomous navigation have been receiving increasing attentions in tandem with the development of micro unmanned aerial vehicles. This paper proposes a kernel cross-correlator (KCC) based algorithm to determine optical flow using a monocular camera, which is named as correlation flow (CF). Correlation flow is able to provide reliable and accurate velocity estimation and is robust to motion blur. In addition, it can also estimate the altitude velocity and yaw rate, which are not available by traditional methods. Autonomous flight tests on a quadcopter show that correlation flow can provide robust trajectory estimation with very low processing power. The source codes are released based on the ROS framework.

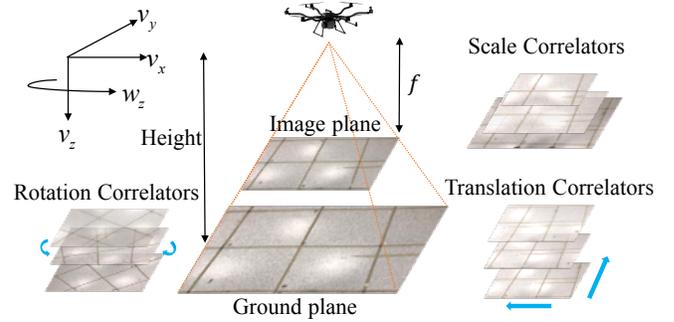

Fig. 1. The framework for camera velocity estimation based on the proposed correlation flow. A downward facing camera mounted on a quadcopter moves arbitrarily in the 3-D space. The horizontal and altitude velocities and the yaw angular rate are estimated by the kernel translation correlator (KTC) and the kernel scale-rotation correlator (KSRC), respectively.

## I. INTRODUCTION

To safely fly in cluttered environments, insects rely on optical flow (OF), which is generated by their own displacement relative to the surroundings [1]. Inspired by this, the optical flow based flight capabilities for unmanned aerial vehicles (UAV) have received increasing attentions, including obstacle avoidance, speed maintenance, odometry estimation, altitude regulation, wall following and corridor centring, orientation control, and landing [1]–[4]. For example, based on optical flow, quadcopters can achieve autonomous navigation and collision avoidance in urban or indoor environments [4]. Optical flow has also been combined with simultaneous localization and mapping (SLAM) algorithms to estimate distances from surrounding environment and stabilize the drone [3]. In recent years, some compact and low-power optical flow sensors have been reported [5]–[7] for micro drones. This dramatically reduces the requirements for onboard energy, sensing, and processing capabilities.

However, the existing algorithms heavily rely on features extracted from the input image, which may be noisy or challenging to extract in flight scenarios, especially at high speed. Therefore, a more accurate and robust method for computing optical flow is needed. To this end, we propose a new optical flow method for velocity estimation based on our recently proposed kernel cross-correlator (KCC) [8], which has been proven to be effective for visual object tracking and human activity recognition using wearable devices. As shown in Fig. 1, we propose a kernel translation correlator (KTC) for horizontal velocity estimation. To achieve robust orientation control and landing, we further develop a kernel scale-rotation correlator (KSRC) for altitude and yaw velocity estimation. Compared with existing methods, correlation flow achieves higher accuracy while still possesses a similar level of computational efficiency. Experiments on autonomous flight of a quadcopter demonstrate the robustness of correlation flow.

## II. RELATED WORK

One of the earliest methods for optical flow is the Horn-Schunck algorithm, that assumes the apparent velocity of the brightness pattern to vary smoothly almost everywhere in the image [9]. By approximating each neighborhood of two consecutive frames using quadratic polynomials, Farneback method estimates the displacement fields from the polynomial expansion coefficients [10]. Using phase correlations, [11] proposes to compute optical flow by block matching, followed by an additional optimization procedure to find smoother motion fields among several candidates. Although these methods yield a high density of flow vectors, they require complex calculation and are sensitive to noise.

To mitigate noise effect, several methods based on feature tracking, e.g. Shi-Tomasi [12], FAST [13], and Lucas-Kanade [14], have been widely used to compute optical flow. Nevertheless, those methods still cannot satisfy the real-time requirements for micro drones [6]. Leveraging on the efficiency of parallel computing, [15] presents an FPGA-based platform for computing the metric optical flow. PX4Flow [5] is an open source and open hardware optical flow sensor using a CMOS camera. It is based on the sum of absolute differences (SAD) block matching algorithm, where the position of the best match in the search area is selected as the resulting flow value. This optical flow sensor is easy to use but sensitive to illumination and motion blur. In [6], the local translation flow is calculated by matching the edge histograms that are obtained by the summation of image gradients in two orthogonal directions. Based on this work, [7] proposes to combine stereo vision and optical flow to estimate velocity and depth for pocket drones.

Deep learning based optical flow methods have also been extensively studied. For example, a thresholded loss for Siamese networks is proposed in [16], where the robustness of trained features for patch matching for optical flow is evaluated. An end-to-end learning strategy is demonstrated for optical flow estimation in [17]. By combining a classical spatial-pyramid method with deep learning, SPyNet trains one deep network per level to compute the optical flow [18]. Although the recent trend towards deep learning based methods mitigates estimation error to an extent, the computation speed is still too low and the cost of training data acquisition is too high for real-time robotic applications, especially micro drones. This opens space for learning techniques that achieve higher accuracy and yield faster training.

Inspired by the fact that local flows are averaged to obtain stable velocity estimation [5], [6], [15], we argue that it may be faster and more robust to predict global flow directly based on learning techniques. To this end, we propose to learn the kernel translation correlator (KTC) to estimate the translation flow. A kernel scale-rotation correlator (KSRC) is further developed to efficiently estimate the scale and rotation flow. Extensive experiments show that correlation flow demonstrates the superiority on accuracy, while still has a similar computational efficiency with traditional algorithms.

## III. PRELIMINARY

In this section we briefly present the definition for kernel cross-correlator on single kernel and single training sample. With respect to [8], we represent signals as 2-D matrices or images, i.e. $\mathbf{z}, \mathbf{x} \in \mathbb{R}^{n_x \times n_y}$, where $\mathbf{z}$ and $\mathbf{x}$ are regarded as the previous and current frame, respectively. The convolution theorem states that cross-correlation becomes element-wise conjugate multiplication in frequency domain. Denote the 2-D fast Fourier transform (FFT) $\mathcal{F}: \mathbb{C}^{n_x \times n_y} \mapsto \mathbb{C}^{n_x \times n_y}$ as $\hat{\cdot}$, so that the cross-correlation of two images $\mathbf{g} = \mathbf{x} * \mathbf{h}$ is equivalent to $\hat{\mathbf{g}} = \hat{\mathbf{x}} \odot \hat{\mathbf{h}}^*$, where the operator $\odot$ and superscript $*$ denote the element-wise multiplication and complex conjugate, respectively. The correlation output $\hat{\mathbf{g}}$ can be transformed back into spatial domain using the inverse FFT. Therefore, the bottleneck of cross-correlation is to compute the forward and backward FFTs, and the complexity of the entire process has an upper bound $\mathcal{O}(N \log N)$, where $N = n_x \times n_y$. Denote the kernel function as $\kappa(\cdot, \cdot)$, such that $\kappa(\mathbf{x}, \mathbf{z}) \in \mathbb{R}$. Given a desired output $\mathbf{g} \in \mathbb{R}^{m_x \times m_y}$, the kernel cross-correlator is defined as:

$$\hat{\mathbf{g}} = \hat{\boldsymbol{\kappa}}_{\mathbf{z}}(\mathbf{x}) \odot \hat{\mathbf{h}}^*, \quad (1)$$

where $\boldsymbol{\kappa}_{\mathbf{z}}(\mathbf{x}) \in \mathbb{R}^{m_x \times m_y}$ is a kernel matrix, with element in the $i_{th}$ row $j_{th}$ column as $\kappa(\mathbf{x}, \mathbf{z}_{ij})$, where $\mathbf{z}_{ij} \in \mathcal{T}(\mathbf{z}) \in \mathbb{R}^{n_x \times n_y}$ is generated from the previous frame $\mathbf{z}$. The transform function $\mathcal{T}(\cdot)$ is predefined for different objectives.

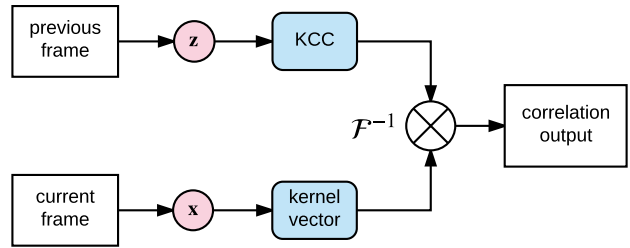

Fig. 2. The computation structure of correlation flow. The position of the maximum response indicates the transformation of images. Except for the FFT, all the operations are element-wise, resulting in efficient computations. Due to the introduction of kernels, correlation flow is robust to noises, motion blur, and image distortions.

The filter $\mathbf{h}$ that maps $\mathbf{z}$ to the desired output $\mathbf{g}$ is to be trained by minimizing the sum of squared errors (SSE) between the kernel cross-correlator and the desired output. To be efficient, we conduct the training in Fourier domain to take advantage of the simple element-wise operation:

$$\min_{\hat{\mathbf{h}}^*} \|\hat{\boldsymbol{\kappa}}_{\mathbf{z}}(\mathbf{z}) \odot \hat{\mathbf{h}}^* - \hat{\mathbf{g}}\|^2 + \lambda \|\hat{\mathbf{h}}^*\|^2, \quad (2)$$

where the second term in (2) is a regularization to prevent overfitting. To solve it we set the first derivative to zero, i.e.,

$$\frac{\partial}{\partial \hat{\mathbf{h}}^*} \left( \|\hat{\boldsymbol{\kappa}}_{\mathbf{z}}(\mathbf{z}) \odot \hat{\mathbf{h}}^* - \hat{\mathbf{g}}\|^2 + \lambda \|\hat{\mathbf{h}}^*\|^2 \right) = 0. \quad (3)$$

Since all the operations in (3) are performed in an element-wise manner, we can obtain a closed-form solution for $\hat{\mathbf{h}}^*$:

$$\hat{\mathbf{h}}^* = \frac{\hat{\mathbf{g}}}{\hat{\boldsymbol{\kappa}}_{\mathbf{z}}(\mathbf{z}) + \lambda}, \quad (4)$$

where the operator $\dot{\div}$ denotes the element-wise division. This solution generates a KCC using a single training sample and a single kernel. One of the advantages of KCC is that any training data $\mathbf{z}$, affine transformation function $\mathcal{T}(\cdot)$, and kernel function $\kappa(\cdot, \cdot)$ can be applied, so that the KCC can be customized for specific applications. The proposed KCC is in contrast with the correlation filter proposed in [19], which only supports training data with circulant structure and non-weighted kernel functions. Readers may refer to [8] for more details.

## IV. CORRELATION FLOW

In this section we present that the translation, scale-rotation flow can be computed by specifying the function $\mathcal{T}(\cdot)$ as translation, scale-rotation transforms, respectively.

### A. Translation Flow

Fig. 2 illustrates the computation structure, in which each prediction takes the previous and current frame as the training and test sample, respectively. To predict the translation flow, the translation transform $\mathcal{T}_T(\cdot)$ on 2-D matrix is applied to generate $\mathbf{z}_{ij}$. Since $\mathbf{z} \in \mathbb{R}^{n_x \times n_y}$, the number of all possible translational shifts $|\mathcal{T}_T(\mathbf{z})| = n_x n_y$, where the operator $|\cdot|$ returns the number of element in a set, and $\mathcal{T}_T(\mathbf{z})$ is the set consisting of all translational shifts of $\mathbf{z}$. Therefore, the size of $\mathbf{h}$ and the kernel matrix $\boldsymbol{\kappa}_{\mathbf{z}}(\mathbf{x})$ equals the size of the image, i.e. $n_x = m_x$, $n_y = m_y$. Without loss of generality, consider the radial basis function (5):

$$\kappa(\mathbf{x}, \mathbf{z}_{ij}) = h\left(\|\mathbf{x} - \mathbf{z}_{ij}\|^2\right). \quad (5)$$

Since the complexity of calculating (5) is $\mathcal{O}(N)$, where $N = n_x n_y$, the complexity of computing a kernel matrix $\boldsymbol{\kappa_z}(\mathbf{x})$ is $\mathcal{O}(N^2)$, which might be infeasible for embedded systems. Fortunately, we find that the kernel matrix can be computed in Fourier domain with complexity $\mathcal{O}(N \log N)$. Firstly, we expand the norm in (5) as:

$$\kappa(\mathbf{x}, \mathbf{z}_{ij}) = h\left(\|\mathbf{x}\|^2 + \|\mathbf{z}_{ij}\|^2 - 2 \cdot \text{Tr}(\mathbf{x}^T \mathbf{z}_i)\right), \quad (6)$$

where the operator $\text{Tr}(\cdot)$ returns the trace of a square matrix. Since $\|\mathbf{x}\|^2$ and $\|\mathbf{z}_{ij}\|^2$ are constants, the kernel matrix can be expressed as:

$$\boldsymbol{\kappa_z}(\mathbf{x}) = h\left(\|\mathbf{x}\|^2 + \|\mathbf{z}\|^2 - 2\left[\text{Tr}(\mathbf{x}^T \mathbf{z}_{ij})\right]_{n_x n_y}\right), \quad (7)$$

where the trace matrix $\left[\text{Tr}(\mathbf{x}^T \mathbf{z}_{ij})\right]_{n_x n_y}$ is defined as:

$$\left[\text{Tr}(\mathbf{x}^T \mathbf{z}_{ij})\right]_{n_x n_y} := \begin{bmatrix} \text{Tr}(\mathbf{x}^T \mathbf{z}_{11}) & \cdots & \text{Tr}(\mathbf{x}^T \mathbf{z}_{1 n_y}) \\ \vdots & \ddots & \vdots \\ \text{Tr}(\mathbf{x}^T \mathbf{z}_{n_x 1}) & \cdots & \text{Tr}(\mathbf{x}^T \mathbf{z}_{n_x n_y}) \end{bmatrix}. \quad (8)$$

From the 2-D correlation theory, $\mathbf{x} * \mathbf{z} = \left[\text{Tr}(\mathbf{x}^T \mathbf{z}_{ij})\right]_{n_x n_y}$. Substituting this into (7), we can obtain

$$\boldsymbol{\kappa_z}(\mathbf{x}) = h\left(\|\mathbf{x}\|^2 + \|\mathbf{z}\|^2 - 2 \cdot \mathbf{x} * \mathbf{z}\right) \quad (9a)$$
$$= h\left(\|\mathbf{x}\|^2 + \|\mathbf{z}\|^2 - 2 \cdot \mathcal{F}^{-1}(\hat{\mathbf{x}} \odot \hat{\mathbf{z}}^*)\right). \quad (9b)$$

The bottleneck of (9b) is the forward and backward FFTs, so that the kernel matrix can be calculated in complexity $\mathcal{O}(N \log N)$. For implementation purpose, the matrix norm in (9b) can be obtained in frequency domain using Parseval's theorem, so that there is no need to store the original signals.

$$\boldsymbol{\kappa_z}(\mathbf{x}) = h\left((\|\hat{\mathbf{x}}\|^2 + \|\hat{\mathbf{z}}\|^2)/N - 2\mathcal{F}^{-1}(\hat{\mathbf{x}} \odot \hat{\mathbf{z}}^*)\right). \quad (10)$$

Based on (10), it is not necessary to generate the sample-based matrices $\mathbf{z}_{ij}$ explicitly, which decreases both space and time complexity dramatically.

The kernel translation correlator (KTC) in 2-D case can then be obtained using (4) and (10). In the experiments, only the center of the desired output $\mathbf{g}$ is set as 1, while all the other positions are set to 0. Intuitively, due to image noise and distortion, it is not possible to obtain an exact single peak for the test sample. Instead, the position of the maximum value in the output is used to find the translation of the test sample. Specifically, the translation $(x_n, y_n)$ of the current frame relative to the training frame is obtained in (11), which is the position of the maximum value in the correlation output relative to the image center $\left(\frac{n_x}{2}, \frac{n_y}{2}\right)$.

$$(x_n, y_n) = \arg\max_{(i,j)} \mathcal{F}^{-1}_{(i,j)}\left(\hat{\boldsymbol{\kappa}}_{\mathbf{z}}(\mathbf{x}) \odot \hat{\mathbf{h}}^*\right) - \left(\frac{n_x}{2}, \frac{n_y}{2}\right), \quad (11)$$

where $\mathcal{F}^{-1}_{(i,j)}(\cdot)$ is an element of the inverse FFT with index $(i, j)$. Therefore, the estimated horizontal metric velocity $(v_x, v_y)$ can be calculated as:

$$(v_x, v_y) = -\frac{h}{\Delta t}\left(\frac{x_n}{f_x}, \frac{x_n}{f_y}\right), \quad (12)$$

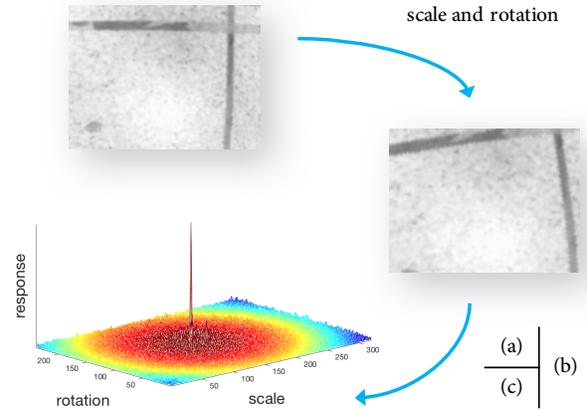

Fig. 3. An example of the KSRC. (a) previous image captured at 30Hz; (b) current image; (c) the response of KSRC. The location of the maximum response relative to the center indicates the transformation pattern. In this example, the image is rotated by 9 degrees and scaled with factor 1.2516.

where $h$ is the height measurement that can be obtained from an altimeter, and $\Delta t$ is the time instant difference between the previous and the current image. $f_x$ and $f_y$ are the focal lengths in $x$ and $y$ direction, respectively.

*B. Scale and Rotation Flow*

As shown in Fig. 1, the altitude velocity $v_z$ and yaw rate $\omega_z$ can be measured by the image scale and rotation transformations, respectively, using a downward facing camera. Similar to the 2-D KTC, it is possible to estimate the scale and rotation using KCC by defining the function $\mathcal{T}(\cdot)$ as scale and rotation transformations, respectively. However, we will show that the complexity is too high for separately calculating the scale and rotation correlators. To accelerate the computation, we propose the kernel scale-rotation correlator (KSRC), which is able to simultaneously estimate the scale and rotation transformations.

In real applications, rotation and scale transformations are usually discretized with specific resolution. Let $\mathbf{z}_{ij}$ be the transformation of $\mathbf{z}$ with specific scale factor $s_i \in \Omega_s$ and rotation angle $\theta_j \in \Omega_\theta$, where $\Omega_s$ and $\Omega_\theta$ are the sets of scale factors and rotation angles, respectively. Let $m_x = |\Omega_s|$ and $m_y = |\Omega_\theta|$, then calculating the scale and rotation kernel vectors are of complexity $\mathcal{O}(m_x N)$ and $\mathcal{O}(m_y N)$, respectively. Therefore, the complexity of (4) is $\mathcal{O}(N \log N + m_{x,y} N)$, which is still bounded by the FFT, especially when $m_x$ or $m_y$ is small. However, if scale and rotation are both present in the image, the complexity of (4) becomes $\mathcal{O}(N \log N + MN)$, where $M = m_x m_y$. This means that the calculation is bounded by the complexity of the kernel matrix with $\mathcal{O}(MN)$, which is very difficult to be carried out in real-time.

To solve this problem, we propose the KSRC as a faster method to calculate the kernel matrix. Since pixels on image boundary are often meaningless and discarded due to the scale and rotation transformations, the pixels near to the image center should account for a greater proportion in the kernel function. Without loss of generality, consider a

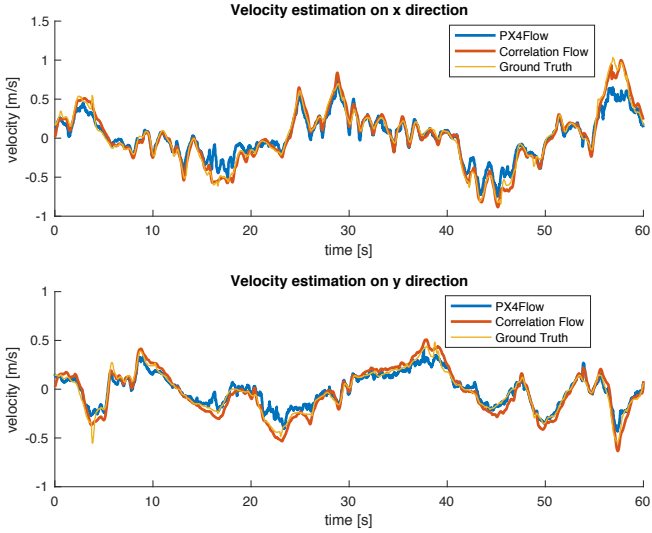

Fig. 4. One example of the velocity estimation from the first 60 seconds of a flight test. It is obvious that the estimation from correlation flow is smoother and more accurate than PX4Flow.

weighted radial basis kernel in (13):

$$\kappa(\mathbf{x}, \mathbf{z}_{ij}) = h\left(\|\mathbf{w} \odot (\mathbf{x} - \mathbf{z}_{ij})\|^2\right), \quad (13)$$

where $\mathbf{w} \in \mathbb{R}_+^{n_x \times n_y}$, **s.t.** $\|\mathbf{w}\|_1 = n_x n_y$ is the weight matrix. Since (13) is weighted, the trace matrix cannot be replaced by the cross-correlation as used in (9). Therefore, the complexity of calculating the kernel matrix $\boldsymbol{\kappa}_\mathbf{z}(\mathbf{x})$ is dominated by the trace matrix, which is still $\mathcal{O}(MN)$. However, we find that this can be solved efficiently using a mapping function, which is defined as $\mathcal{M} : \mathbb{R}^{n_x \times n_y} \mapsto \mathbb{R}^{m_x \times m_y}$ in (14). For simplicity, we denote it as $\tilde{\cdot}$, i.e.,

$$\tilde{\mathbf{x}} = \mathcal{M}(\mathbf{x}), \quad \text{s.t.} \quad \text{Tr}(\tilde{\mathbf{x}}^T \tilde{\mathbf{y}}) = \text{Tr}((\mathbf{w} \odot \mathbf{x})^T (\mathbf{w} \odot \mathbf{y})), \quad (14)$$

where $\mathbf{x}, \mathbf{y} \in \mathbb{R}^{n_x \times n_y}$. Since $\|\mathbf{x}\|^2 = \text{Tr}(\mathbf{x}^T \mathbf{x})$, $\|\tilde{\mathbf{x}}\|^2 = \|\mathbf{w} \odot \mathbf{x}\|^2$, substituting (14) into (13), we have

$$\kappa(\mathbf{x}, \mathbf{z}_{ij}) = h\left(\|\tilde{\mathbf{x}}\|^2 + \|\tilde{\mathbf{z}}_{ij}\|^2 - 2 \cdot \text{Tr}(\tilde{\mathbf{x}}^T \tilde{\mathbf{z}}_{ij})\right), \quad (15)$$

where $\tilde{\mathbf{z}}_{ij} = \mathcal{M}(\mathbf{z}_{ij})$. Therefore, the kernel matrix can be calculated as:

$$\boldsymbol{\kappa}_\mathbf{z}(\mathbf{x}) = h\left(\|\tilde{\mathbf{x}}\|^2 + \|\tilde{\mathbf{z}}\|^2 - 2\left[\text{Tr}(\tilde{\mathbf{x}}^T \tilde{\mathbf{z}}_{ij})\right]_{m_x m_y}\right), \quad (16)$$

where the trace matrix $\left[\text{Tr}(\tilde{\mathbf{x}}^T \tilde{\mathbf{z}}_{ij})\right]_{m_x m_y}$ is defined similarly to (8). Inspired by the Fourier-Mellin transform [20], the trace matrix can be converted to cross-correlation, if the element of the weight matrix $\mathbf{w}$ is set as (17):

$$\mathbf{w}_{[x,y]} \propto \left|\left\{(i,j) \middle| \begin{matrix} x = \langle \exp \xi_i \cos \theta_j \rangle \\ y = \langle \exp \xi_i \sin \theta_j \rangle \end{matrix}\right\}\right|, \quad (17)$$

where $(x, y)$ are image coordinates relative to the image center and the operator $\langle \cdot \rangle$ returns the nearest integer of a real number. In this sense, $\mathcal{M}(\cdot)$ becomes the log-polar transform with coordinates $(\xi, \theta)$, where $\xi = \log \sqrt{x^2 + y^2}$ and $\theta = \text{atan2}(y, x)$. Therefore, the image mapping $\tilde{\mathbf{z}}$ in the log-polar plane satisfies (18), which means that $\mathbf{z}_{ij}$ is the transformation of $\mathbf{z}$ with scale factor $s_i = \exp \xi_i$ and rotation angle $\theta_j$.

$$\tilde{\mathbf{z}}_{ij}(\xi, \theta) = \tilde{\mathbf{z}}(\xi - \xi_i, \theta - \theta_j). \quad (18)$$

It is easy to verify that the weight matrix $\mathbf{w}$ defined in (17) satisfies the intuitive idea that pixels near to the center weigh more than those near to the boundary. Substituting (18) into (16) and ignoring the boundary effect, we can approximate the trace matrix in (16) by cross-correlation in (19a) and element-wise multiplication in (19b).

$$\boldsymbol{\kappa}_\mathbf{z}(\mathbf{x}) = h\left(\|\tilde{\mathbf{x}}\|^2 + \|\tilde{\mathbf{z}}\|^2 - 2 \cdot \tilde{\mathbf{x}} * \tilde{\mathbf{z}}\right) \quad (19a)$$

$$= h\left((\|\hat{\tilde{\mathbf{x}}}\|^2 + \|\hat{\tilde{\mathbf{z}}}\|^2)/M - 2\mathcal{F}^{-1}(\hat{\tilde{\mathbf{x}}} \odot \hat{\tilde{\mathbf{z}}})\right). \quad (19b)$$

Dominated by the forward and backward FFTs in (19b), the complexity of calculating the kernel matrix is reduced to $\mathcal{O}(M \log M + M)$, which is much smaller than $\mathcal{O}(N \log N + MN)$. An example of KSRC is shown in Fig. 3. Assuming that $(x_m, y_m)$ is the translation of the maximum value in the correlation output, which is obtained similarly to (11), we can compute the altitude velocity $v_z$ and yaw rate $\omega_z$ as:

$$v_z = \frac{(s-1)h}{\Delta t}, \quad \omega_z = \frac{2\pi y_m}{m_y \Delta t}, \quad (20)$$

where $s$ is the estimated scale factor:

$$s = \exp\left(\frac{\log(m_y/2)}{m_x} x_m\right). \quad (21)$$

Note that we can also obtain the altitude velocity by differentiating the altimeter measurements, but it is very noisy. One possible solution is to fuse the two sources of information, however, it is out of the scope of this paper.

## V. EXPERIMENTS

Successful autonomous navigation of drones depends on a robust optical flow system, which is used to provide accurate velocity estimation, and hence improve the position estimation [2], [3], [5], [6]. In this section, extensive experiments on velocity estimation, autonomous flight, and battery life hovering test demonstrate the superior performance of the proposed **correlation flow** system.

### A. Implementation

*1) Software:* Since $\mathbf{g}$ is not changed during the training stage, $\hat{\mathbf{g}}$ only needs to be calculated once when starting the program. The regularization parameter $\lambda$ in (4) is set as 0.1 to prevent overfitting. The Gaussian kernel (22), which is proved to be robust to noise and distortion is used:

$$\kappa(\mathbf{x}, \mathbf{z}) = \exp\left(-\frac{\|\mathbf{x} - \mathbf{z}\|^2}{2\sigma^2}\right), \qquad (22)$$

where $\sigma$ is set as $0.2$. In the experiments, we found that these parameters are not sensitive to the test environments, since the results are not much affected by different choices of these parameters in different test scenarios. Note that to obtain a higher update rate, the optical flow systems in [5] and [6] only process gray scale images with size $64 \times 64$ and $128 \times 96$, respectively. However, because of the high efficiency of FFT and element-wise operation, our correlation flow is able to process images with size $320 \times 240$, resulting in much higher flow resolution, yet still with high update rate (real-time) on an ultra-low-power processor. We implement and test our framework on Ubuntu with robot operating system (ROS). The source codes are released at https://github.com/wang-chen/correlation_flow.

TABLE I
COMPARISON ON HORIZONTAL VELOCITY WITH PX4FLOW. (m/s)

| Test | Correlation Flow | | PX4Flow | |
|------|-------|-------|-------|-------|
|      | RMSE  | MAE   | RMSE  | MAE   |
| 01   | **0.069** | **0.054** | 0.145 | 0.117 |
| 02   | **0.072** | **0.058** | 0.167 | 0.138 |
| 03   | **0.074** | **0.057** | 0.148 | 0.123 |
| 04   | **0.069** | **0.053** | 0.140 | 0.110 |
| 05   | **0.076** | **0.064** | 0.132 | 0.100 |
| Mean | **0.072** | **0.057** | 0.146 | 0.118 |

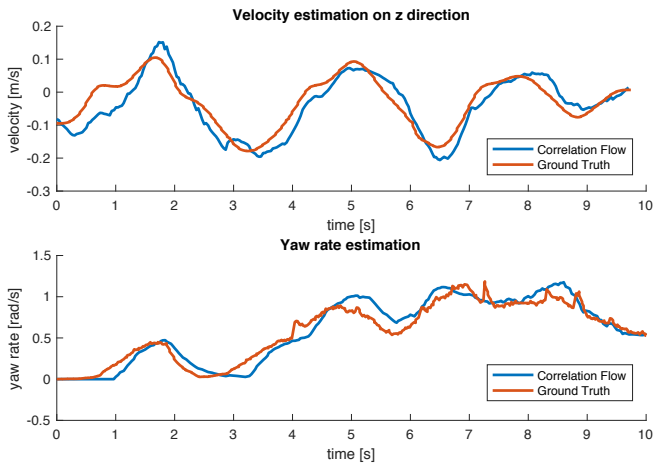

Fig. 5. The estimated yaw rate and altitude velocity compared with ground truth from Vicon system.

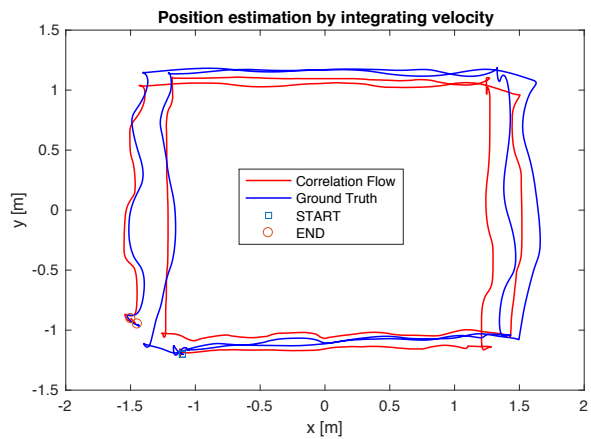

Fig. 6. The estimated position by integrating the velocity estimation during fully autonomous flight. The drone tries to fly along rectangular paths.

*2) Platform:* Limited by the payloads and power consumption, we choose a credit card-sized computing board, UP, that is equipped with an ultra-low-power processor x5-Z8350 with scenario design power of only 2W. Tests conducted on this computing board show that correlation flow leaves enough computational resources for other tasks, such as localization, path planning, graph optimization [21], and Non-Iterative SLAM [22]. Together with the computing board, an industrial IDS uEye UV-1551LE CMOS camera is mounted ventrally on a micro-quadcopter as shown in Fig. 7. All the experiments are performed in a Vicon-equipped room, that can provide very accurate pose and velocity estimation at 50Hz as the ground truth. The experimental results are recorded on-board during fully autonomous flight.

*B. Velocity Estimation and Comparison*

The velocity estimation is compared with one of the state-of-the-art methods, PX4Flow [5], which integrates a gyroscope and a sonar altimeter. It might be one of the most suitable methods to compare, since it has been commercialized, fully tuned and tested by the robotic community. To be fair, it is mounted on the same platform mentioned in Section V-A.2. Limited by the bandwidth of serial port, the images from PX4Flow cannot be recorded with full rate. Hence, only the estimation results are saved for comparison. Five on-board flight tests, each of which lasts more than 2 minutes, are performed in order to cover distinct traveling distances, speeds, dynamics, and illumination conditions. Both PX4Flow and correlation flow have an update rate of 30Hz. Table I shows the accuracy comparison in terms of root mean squared error (RMSE) and median absolute error (MAE). It is obvious that correlation flow **outperforms** PX4Flow in every flight test, resulting in improving the accuracy (RMSE) by more than $100\%$. Fig. 4 shows the plot of velocity estimation from one of the flight tests. It can be seen that correlation flow is able to provide more accurate and smoother velocity estimation than PX4Flow.

Since most of the existing methods, including [5] and [6], are unable to provide the estimation on altitude and yaw change, we only compare the scale and rotation flow with the ground truth from the Vicon system. Fig. 5 presents an example of the estimation results of the scale and rotation flow. Note that the ground truth of yaw rate is obtained by differentiating the attitude estimation from Vicon since it cannot estimate yaw rate directly. The altitude and yaw velocity estimation is crucial for robust orientation control and landing, but this is out of the scope of this paper.

*C. Autonomous Flight*

The main objective of this section is to show the potential of correlation flow for fully autonomous flight. The estimation from correlation flow can be fused by the state estimator in the flight controller based on an extended Kalman filter. Hence, we can estimate the position by integrating the estimated velocity. Fig. 7 illustrates this simple control scheme. The velocity command is sent to the controller, for which the quadcopter tries to follow a simple rectangular trajectory. Fig. 6 presents the estimated trajectory which is obtained in real-time flight lasting about 127s. It can be seen that fully autonomous flight is enabled without other positional device, except for an inertial measurement unit (IMU) in the controller. The mean trajectory estimation error is about $0.085$m with standard deviation of $0.051$m, which is acceptable for most of the flight applications. This demonstrates the feasibility of correlation flow for autonomous flight.

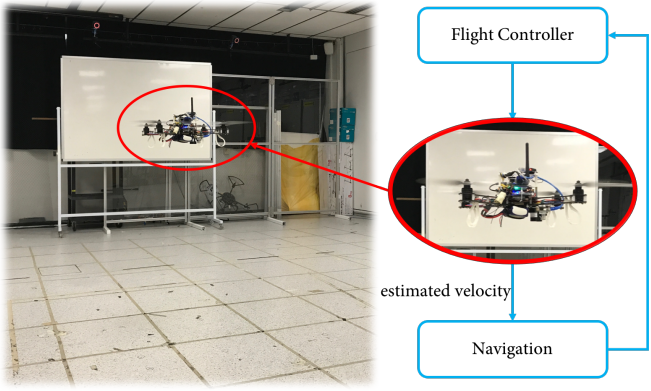

Fig. 7. The flight environment and the control scheme.

*D. Hovering Test*

As the same as all the other optical flow methods, the position estimation of correlation flow will also drift due to long-term integration. This section presents the limit test for drifting, in which the maximum hovering time indicates the drifting speed and is measured for demonstration. The timer is stopped if the quadcopter drifts too much when the autonomous mode is switched back to manual control. Testing environment is shown in Fig. 7, which is a screenshot during the flight. Limited by the payloads including the sensors, the battery life of the platform mentioned in Section V-A.2 is about $5$ min. The quadcopter can hover within the flight area during the whole battery life, which further demonstrates the robustness of correlation flow.

## VI. CONCLUSION

In this paper we propose a robust and computationally efficient optical flow method, called correlation flow for robot velocity estimation using a monocular camera. We introduce a kernel translation correlator and a kernel scale-rotation correlator for the camera motion prediction. Due to the high efficiency of fast Fourier transform, our method is able to run in real-time on an ultra-low-power processor. Experiments on velocity estimation show that correlation flow provides more reliable results than PX4Flow. Autonomous flight and hovering tests demonstrate that correlation flow is able to provide robust trajectory estimation at very low computational cost. The source codes are released for research purpose.

## ACKNOWLEDGMENTS

The authors would like to thank Mr. *Junjun Wang*, Hoang Minh-Chung, and Xu Fang for their help in the experiments. This research was partially supported by the ST Engineering-NTU Corporate Lab funded by the NRF Singapore.